\documentclass[sigconf]{acmart}

\definecolor{blue}{HTML}{1F77B4}
\definecolor{orange}{HTML}{FF7F0E}
\definecolor{green}{HTML}{2CA02C}

\AtBeginDocument{%
  \providecommand\BibTeX{{%
    \normalfont B\kern-0.5em{\scshape i\kern-0.25em b}\kern-0.8em\TeX}}}

\usepackage{amsmath}
\usepackage{amsfonts}
\usepackage{lipsum,booktabs}
\usepackage{algorithm}
\usepackage{algorithmic}
\usepackage{booktabs} 
\usepackage{caption} 
\usepackage{subcaption} 
\usepackage{graphicx}
\usepackage{pgfplots}
\usepackage{makecell}
\usepackage[all]{nowidow}
\usepackage[utf8]{inputenc}
\usepackage{todonotes}
\usepackage{tikz}
\usetikzlibrary{er,positioning,bayesnet}
\usepackage{multicol}
\usepackage{hyperref}
\usepackage[inline]{enumitem} 

\copyrightyear{2022}
\acmYear{2022}
\setcopyright{acmlicensed}\acmConference[BCB '22]{13th ACM International Conference on Bioinformatics, Computational Biology and Health Informatics}{August 7--10, 2022}{Northbrook, IL, USA}
\acmBooktitle{13th ACM International Conference on Bioinformatics, Computational Biology and Health Informatics (BCB '22), August 7--10, 2022, Northbrook, IL, USA}
\acmPrice{15.00}
\acmDOI{10.1145/3535508.3545538}
\acmISBN{978-1-4503-9386-7/22/08}
\begin{document}

\title{Transparent Single-Cell Set Classification with Kernel Mean Embeddings}

\author{Siyuan Shan}
\email{siyuanshan@cs.unc.edu}
\affiliation{%
  \institution{Department of Computer Science, UNC-Chapel Hill
}
\country{USA}
}

\author{Vishal Athreya Baskaran}
\email{athreya@cs.unc.edu}
\affiliation{%
  \institution{Department of Computer Science, UNC-Chapel Hill
}
\country{USA}
}

\author{Haidong Yi}
\email{haidyi@cs.unc.edu}
\affiliation{%
  \institution{Department of Computer Science, UNC-Chapel Hill
}
\country{USA}
}

\author{Jolene Ranek}
\email{ranekj@live.unc.edu}
\affiliation{%
  \institution{Computational Medicine Program, Curriculum in Bioinformatics and Computational Biology, UNC-Chapel Hill
}
\country{USA}
}

\author{Natalie Stanley}
\email{natalies@cs.unc.edu}
\affiliation{%
  \institution{Computational Medicine Program, Department of Computer Science, UNC-Chapel Hill
}
\country{USA}
}

\author{Junier B. Oliva}
\email{joliva@cs.unc.edu}
\affiliation{%
  \institution{Department of Computer Science, UNC-Chapel Hill
}
\country{USA}
}

\renewcommand{\shortauthors}{Shan et al.}

\begin{abstract}
  Modern single-cell flow and mass cytometry technologies measure the expression of several proteins of the individual cells within a blood or tissue sample. Each profiled biological sample is thus represented by a set of hundreds of thousands of multidimensional cell feature vectors, which incurs a high computational cost to predict each biological sample's associated phenotype with machine learning models. Such a large set cardinality also limits the interpretability of machine learning models due to the difficulty in tracking how each individual cell influences the ultimate prediction. We propose using Kernel Mean Embedding to encode the cellular landscape of each profiled biological sample. Although our foremost goal is to make a more transparent model, we find that our method achieves comparable or better accuracies than the state-of-the-art gating-free methods through a simple linear classifier.  As a result, our model contains few parameters but still performs similarly to deep learning models with millions of parameters. In contrast with deep learning approaches, the linearity and sub-selection step of our model makes it easy to interpret  classification results. Analysis further shows that our method admits rich biological interpretability for linking cellular heterogeneity to clinical phenotype.

\end{abstract}

\ccsdesc[300]{Computing methodologies~Machine learning algorithms}
\ccsdesc[500]{Applied computing~Computational biology}

\keywords{Cytometry, Single-Cell Bioinformatics , Clinical Prediction, Flow Cytometry, Mass Cytometry, Kernel Methods}


\maketitle
\section{Introduction}
\label{sec:intro}


Modern immune profiling techniques such as flow and mass cytometry (CyTOF) enable comprehensive profiling of immunological heterogeneity across a multi-patient cohort \cite{markdavis,perplexed}. In recent years, such technologies have been applied to numerous clinical applications. In particular, these assays allow for both the phenotypic and functional characterization of immune cells based on the simultaneous measurement of 10-45 protein markers \cite{bendallCytometry}. To further connect the diversity, abundance, and functional state of specific immune cell-types to clinical outcomes or external variables, modern bioinformatics approaches have focused on how to engineer or learn a set of ``immune features'' that adequately encodes a profiled individual's immunological landscape. In general, existing approaches for creating immunological features either rely heavily on manual human effort \cite{pree,ganio,perplexed}, clustering \cite{citrus,spade,vopo}, or do not produce immunological features that are readily interpretable or informative for follow-up experiments, diagnostics, or treatment strategies \cite{yi2021cytoset}. To efficiently and accurately link cellular heterogeneity to clinical outcomes or external variables, here we introduce 
Cell Kernel Mean Embedding (CKME), a method based on kernel mean embeddings \cite{muandet2017kernel} that directly featurizes cells in samples according to their aggregate distribution. 
We show that CKME is simple enough to be readily interpreted by a human, and learns individual cell scores. Empirical studies show that CKME achieves competitive to state-of-the-art classification accuracy in clinical outcome prediction tasks. \\

%


\begin{figure}[t]
\begin{center}
  \includegraphics[width=0.84\columnwidth]{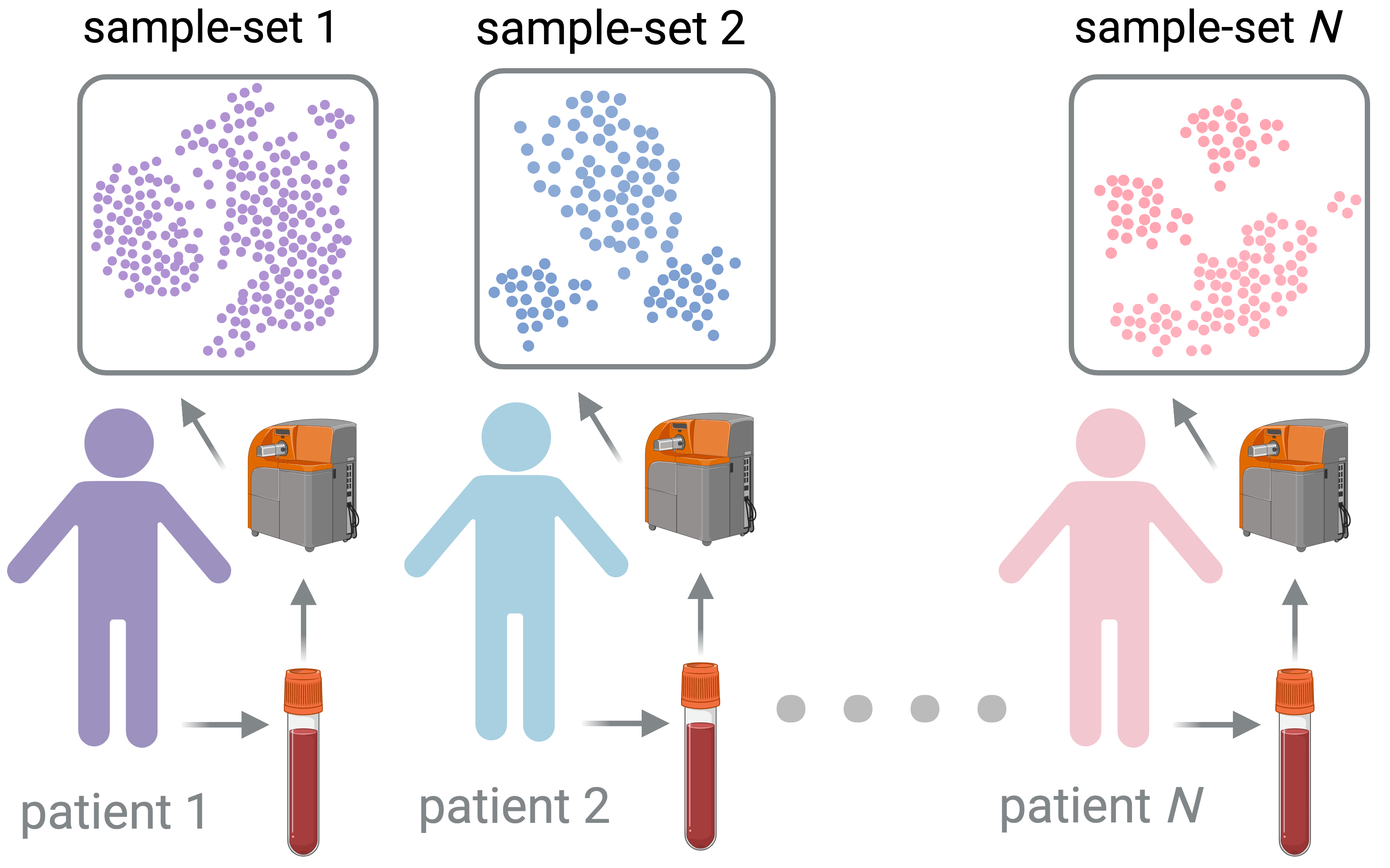}
\caption{Single-cell profiles produce a \emph{sample-set} of (typically thousands of) multidimensional feature vectors of measured features per cell (illustrated as scatter plots). Thus, a dataset of multiple biological samples will result in a dataset of multiple sets (shown above).}
\label{fig:single_cell}
\end{center}  
\end{figure}

\noindent \textbf{Flow and Mass Cytometry Assays Produce a Dataset of Sets} \ \
To comprehensively profile the immune systems of biological samples collected across multiple individuals, a unique data structure of multiple \emph{sample-sets} is ultimately produced. Each sample-set is likely to contain hundreds
of thousands of cells collected from an individual, and machine learning models are often used to predict their associated clinical or experimental labels.  This concept is illustrated in Fig. \ref{fig:single_cell}. In profiling multiple individuals with a single-cell technology, a \textit{dataset of sets} is produced and is a non-traditional data structure that breaks the standard convention of a \textit{dataset of feature vectors}. This produces complexity for learning over individuals (e.g. patients), as single-cell data requires a model to characterize and compare across multiple sets of multiple vectors (where each vector represents a cell).
Another challenge of analyzing single-cell data is the large number of cells in each sample-set, which incurs a computational cost and obscures the decision-making process of machine learning models. \\

\noindent \textbf{Related Work}\ \
The first class of methods for identifying coherent cell-populations and specifying their associated immunological features are \emph{gating-based}\footnote{Gating refers to partitioning cells into populations}, and operate by first assigning cells to populations either manually, or in an automated manner by applying an unsupervised clustering approach \cite{perplexed, flowcap}. After an individual's cells have been assigned to their respective cell-populations, corresponding immunological features are specified by 1) computing frequencies or the proportion of cells assigned to each population and 2) computing functional readouts as the median expression of a particular functional marker \cite{vopo}. Though gating-based approaches can be automatic without human intervention, they have several drawbacks: 1) they tend to be sensitive to variation in the parameters of the underlying clustering algorithms \cite{vopo}. 2) the underlying clustering algorithms typically do not scale well to millions of cells. 3) they typically need to rerun the expensive clustering process whenever a new sample-set is given \cite{Kiselev2019-zx}. Furthermore, partitioning cells into discrete clusters can be inadequate in the context of continuous phenotypes such as drug treatment response and ultimately limits the capacity of having a single-cell resolution of the data.

To address these limitations, the second class of methods are \emph{gating-free} and rely on representing, or making predictions based on individual cells \cite{cellCNN,cytoDX,yi2021cytoset}. 
For example, CytoDx \cite{cytoDX} uses a linear model to first compute the response of every single cell individually, then the responses are aggregated by a mean pooling function. Finally, the aggregated result is used by another linear model to predict clinical outcomes. 
CellCNN \cite{cellCNN} uses a 1-d convolution layer to learn filters that are responsive to marker profiles of cells and aggregates the responses of cells to different filters via a pooling layer for prediction. Since both CytoDx and CellCNN only contain a single linear or convolution layer before the pooling operation, they enable human interpretation. However, this property also limits their expressive power. 
To enhance model flexibility, CytoSet \cite{yi2021cytoset} employs a deep learning model similar to Deep Set \cite{zaheer2017deep} to handle set data with a permutation invariant architecture that stacks multiple intermediate permutation \emph{equivariant} neural network layers.
As a result, the sample-set featurization that CytoSet achieves, while discriminative and accurate, is opaque and a black box, making it difficult to analyze downstream for biological discoveries.
In addition, CytoSet usually contains millions of parameters, further obfuscating the underlying predictive mechanisms. 
Our method CKME belongs to this \textit{gating-free} class; we show that CKME achieves interpretability without harming model expressiveness. \\


\noindent \textbf{Summary and Contribution of Results} \ \ 
CKME tackles the challenges highlighted above by computing a kernel mean embedding (KME) \cite{muandet2017kernel} of sample-sets to represent the characteristics of each sample-set's composition (and distribution).
For classification tasks, we train a linear classifier (e.g. linear SVM) in the KME feature space. Despite its simplicity, CKME achieves comparable to state-of-the-art gating-free performance on three flow and mass cytometry datasets with multiple sample-sets and associated clinical outcomes. 
We show that CKME may be used to obtain an individual cell score for each cell in a sample-set, where the ultimate prediction is the result of the average cell score.
As a result, CKME is highly interpretable since one can quantify the individual contribution of every cell to the final prediction. 
Predictions are then synthesized further using kernel herding \cite{chen2010super}, which identifies key cells to retain in a subsample.
We leverage the transparency of CKME in a thorough analysis to understand predictions.
First, we study the semantics of cell scores from CKME, and show that they follow several desirable, intuitive properties.
After, we explore cell scores to obtain biological insights from our models, and show that discoveries from CKME are consistent with existing knowledge.
We release publicly available code at \url{https://github.com/shansiliu95/CKME}.\\


\section{Methods}
\label{sec:methods}
{\bf Notation for Sample-sets}\ \  A \textit{sample} refers to the collection of cells from an individual's blood or tissue, and is further represented as a \textit{sample-set} of many, $n$, cells: $\mathcal{X}= \{x^{(i)}\}_{i=1}^n$, where $x^{(i)} \in \mathbb{R}^d$ denotes the vector of $d$ features (e.g. proteins or genes) measured in cell $i$. A multi-sample dataset $\mathcal{D}$ contains multiple sample-sets (across multiple, $N$, individuals and conditions): $\mathcal{D}=\big\{\mathcal{X}^{(k)}\big\}_{k=1}^{N} = \big\{\{x^{(k,i)}\}_{i=1}^{n_k}\big\}_{k=1}^{N}$,  where $\mathcal{X}^{(k)}=\{x^{(k,i)}\}_{i=1}^{n_k}$ is the sample-set for the $k$-th profiled biological sample. Sample-sets often have an associated labels of interest, $y$, such as their clinical phenotype or conditions. In this case, our dataset consists of sample-set and label tuples: $\mathcal{D}=\big\{(\mathcal{X}^{(k)}, y^{(k)})\big\}_{k=1}^{N}$.\\

\begin{figure*}[t!]
\begin{center}
  \includegraphics[width=1.5\columnwidth]{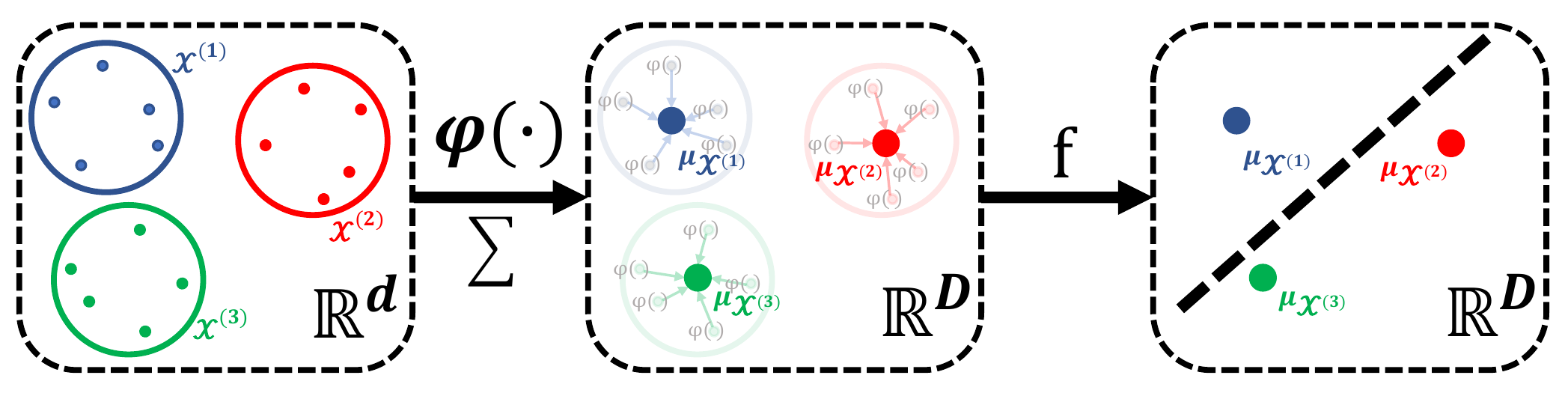}
\caption{The pipeline of CKME to process the dataset and train the classifier. The left rectangle shows a dataset with 3 sample-sets and each sample-set contains 5 features vectors in the original feature space $\mathbb{R}^d$. In the middle rectangle, we use $\varphi(\cdot)$ to transform the data into random Fourier feature space $\mathbb{R}^D$ and compute the mean embedding $\mu_{\mathcal{X}^{(1)}},\mu_{\mathcal{X}^{(2)}},\mu_{\mathcal{X}^{(3)}}$ for every sample-set (eq.~\ref{eq:randfeat_mme}). Finally, in the right rectangle, we train a linear discriminative model $f$ (eq.~\ref{eq:randfeat_mmepred}) upon the mean embeddings to predict the label of each sample-set.} 
\label{fig:framework}
\end{center}  
\end{figure*}

\noindent {\bf Problem Formulation}\ \  Given a dataset of multiple sample-sets as discussed above, we wish to build a discriminative model 
to predict patient-level phenotypes or conditions based on the cellular composition that is found within a sample-set, $\mathcal{X} \rightarrow y$. If the model is human interpretable, it will help researchers make sense of how the cell feature vectors in a sample-set influence the predicted phenotype or conditions of a patient, which eventually will lead to an improved understanding of biological phenomena and enable better diagnoses and treatment for patients. Below we propose a methodology to featurize and classify input sample-sets in a way that is more transparent and human-understandable than comparably performant models (e.g.~\cite{yi2021cytoset}).  \\

\noindent {\bf Kernel Mean Embedding}\ \ To classify labels of interest, $y$, given an input sample-set, $\mathcal{X}$, we featurize $\mathcal{X}$ so that we may learn an estimator over those features. However, unlike traditional data-analysis, which featurizes a single vector instance $x \in \mathbb{R}^d$ with features $\phi(x): \mathbb{R}^d \mapsto \mathbb{R}^q$, here we  featurize a \emph{set of multiple} vectors (one vector for each cell in a sample-set) $\mathcal{X}= \{x^{(i)}\}_{i=1}^n$, 
$\phi(\mathcal{X}) \in \mathbb{R}^q$.
Featurizing a set presents a myriad of challenges since typical machine learning approaches are constructed for \emph{statically-sized, ordered} inputs. In contrast, sets are of \emph{varying cardinalities} and are \emph{unordered}. Hence, straight-forward approaches, such as concatenating the sample-set elements into a single vector $(x^{(1)}, \ldots, x^{(n)}) \in \mathbb{R}^{nd}$ shall fail to provide mappings that do not depend on the order that elements appear in. To respect the \emph{unordered} property of sample-sets one must carefully featurize $\mathcal{X}$ in a way that is \emph{permutation-invariant}. That is, the features $\phi(\mathcal{X})$ should be unchanged regardless of what order that the elements of $\mathcal{X}$ are processed. Recently, there have been multiple efforts to create methods based on neural networks to featurize sets in a permutation invariant manner \cite{qi2017pointnet,zaheer2017deep,shi2020deep, li2020exchangeable, shan2021nrtsi}. Although these approaches provide expressive, non-linear, discriminative features, they are often opaque and difficult to understand in how they lead to their ultimate predictions. In contrast, we propose an approach based on kernels and random features that is more transparent and understandable whilst being comparably accurate. 

Kernel methods have achieved great success in many distinct machine learning tasks, including: classification \cite{cortes1995support}, regression \cite{vovk2013kernel}, and dimensionality reduction \cite{mika1998kernel}. They utilize a positive definite kernel function $\texttt{k} : \mathbb{R}^d \times \mathbb{R}^d \mapsto \mathbb{R}\,$\,\footnote{Note that kernels may be defined over non-real domains, this is omitted for simplicity.}, which induces a reproducing kernel Hilbert space (RKHS) (e.g.~see \cite{berlinet2011reproducing} for further details). Kernels have also been deployed for representing a distribution, $p$, with the \emph{kernel mean embedding} $\mu_p: \mathbb{R}^d \mapsto \mathbb{R}$:
\begin{equation}
    \mu_p(\cdot) \equiv \mathbb{E}_{x\sim p}[\texttt{k}(x, \cdot)].
\end{equation}
Note that $\mu_p$ is itself a function (see e.g.~\cite{muandet2017kernel} for more details). For ``\emph{characteristic}'' kernels, $\texttt{k}$, such as the common radial-basis function (RBF) kernel $\texttt{k}(x, x^\prime) = \exp(-\frac{1}{2 \gamma} ||x - x^\prime||^2)$, the kernel mean embedding will be unique to its distribution; i.e., for characteristic kernels, $||\mu_p - \mu_q|| = 0$ if and only if $p = q$. In general, the distance\footnote{In the RKHS norm.} $||\mu_p - \mu_q||$ induces a divergence, the maximum mean discrepancy (MMD) \cite{gretton2008kernel}, between distributions.

For our purposes, we propose to use kernel mean embeddings to featurize sample-sets:
\begin{equation}
    \mu_{\mathcal{X}}(\cdot) \equiv \frac{1}{n} \sum_{i=1}^n \texttt{k}(x^{(i)}, \cdot) \approx \mu_p(\cdot), \label{eq:setemb}
\end{equation}
where $p$ is the underlying distribution (of cell features) that $\mathcal{X}$ was sampled from. That is, the set embedding $\mu_{\mathcal{X}}$ (eq.~\ref{eq:setemb}) also approximately embeds the underlying distribution of cells that the sample-set was derived from. To produce a real-valued output from the mean embedding, one would take the (RKHS) inner product with a learned function $f$: 
\begin{equation}
    \langle \mu_{\mathcal{X}}, f \rangle = \frac{1}{n} \sum_{i=1}^n \langle \texttt{k}(x^{(i)}, \cdot), f(\cdot) \rangle = \frac{1}{n} \sum_{i=1}^n  f(x^{(i)}), \label{eq:mmeout}
\end{equation}
where the last term follows from the reproducing property of the RKHS.
For example, eq.~\ref{eq:mmeout} can be used to output the log-odds for a target $y$ given $\mathcal{X}$: $p(y=1 | \mathcal{X}) = (1 + \exp(- \langle \mu_{\mathcal{X}}, f \rangle ))^{-1}$. Using the representer theorem \cite{scholkopf2001generalized} it can be shown that $f$ may be learned and represented using a ``\emph{Gram}'' matrix of pairwise kernel evaluations, $\texttt{k}(x, x^\prime)$. This, however, will be prohibitive in larger datasets. Instead of working directly with a kernel $\texttt{k}$, we propose to leverage random Fourier features for computational efficiency and simplicity.\\

\noindent {\bf Random Fourier Features}\ \  We propose to use random Fourier frequency features \cite{rahimi2007random} to build our mean embedding \cite{muandet2017kernel}. For a shift-invariant kernel (such as the RBF kernel), random Fourier features provide a feature map $\varphi(x) \in \mathbb{R}^D$ such that the dot product in feature space approximates the kernel evaluation, $\varphi(x)^T \varphi(x^\prime) \approx \texttt{k}(x, x^\prime)$. I.e. $\varphi(x)$ acts as an approximate \emph{primal space} for the kernel $\texttt{k}$ (please see \cite{rahimi2007random, sutherland2015error, sutherland2016linear, oliva2014fast} for further details).  Using the dot product of $\varphi(x)$, our mean embedding becomes: 
\begin{equation}
    \mu_\mathcal{X} = \frac{1}{n} \sum_{i=1}^n \varphi(x^{(i)}) \in \mathbb{R}^D, \quad \mu_\mathcal{X}(x^\prime) = \frac{1}{n} \sum_{i=1}^n\varphi(x^{(i)})^T \varphi(x^\prime)
    \label{eq:randfeat_mme}
\end{equation}
where {\small $\varphi(x) = \left(\sin(\omega_1^T x), \ldots, \sin(\omega_{D/2}^T x),\, \cos(\omega_1^T x), \ldots, \cos(\omega_{D/2}^T x) \right)$} with random frequencies $\omega_j \sim  \rho$ drawn once (and subsequently held fixed) from a distribution $\rho$ that depends on the kernel $\texttt{k}$. For instance, for the RBF kernel, $\rho$ is an \emph{iid} multivariate independent normal with mean 0 and variance that depends (inversely) on the bandwidth of the kernel, $\gamma$.

When computing the mean embedding in the $\varphi(\cdot)$ feature space (eq.~\ref{eq:randfeat_mme}), one may directly map $\mu_\mathcal{X}$ to a real value with a dot product with learned coefficient $\beta \in \mathbb{R}^D$:
\begin{equation}
    \mu_\mathcal{X}^T \beta = \left(\frac{1}{n} \sum_{i=1}^n \varphi(x^{(i)}) \right)^T \beta \,= \frac{1}{n}\sum_{i=1}^n \varphi(x^{(i)})^T \beta.
    \label{eq:randfeat_mmeout}
\end{equation}
That is, we may learn a linear model directly operating over the $D$ dimensional feature vectors $\mu_\mathcal{X}$, which are composed of the average random features found in a sample-set. Below we expound on how to build a discriminative model based on $\mu_\mathcal{X}$. \\


\noindent {\bf Linear Classifier with Interpretable Scores}\ \ With the mean embedding of sample-sets $\left\{ \mu_{\mathcal{X}^{(k)}} \right\}_{k=1}^N$, $\mu_{\mathcal{X}^{(k)}} \in \mathbb{R}^D$, we can build a discriminative model $f: \mathbb{R}^D \rightarrow \mathbb{R}$ to predict their labels. If we choose $f$ as a linear model (e.g. linear SVM), then $f(\mu_{\mathcal{X}})$ can be generally expressed as 
\begin{equation}
    f(\mu_{\mathcal{X}^{(k)}}) = \mu_{\mathcal{X}^{(k)}}^T\beta + b = \frac{1}{n_k}\sum_{i=1}^{n_k}\underbrace{\varphi(x^{(k,i)})^T \beta + b}_{s^{(k,i)}} = \frac{1}{n_k}\sum_{i=1}^{n_k} s^{(k,i)},
    \label{eq:randfeat_mmepred}
\end{equation}
where $\beta \in \mathbb{R}^D$ and $b \in \mathbb{R}$ are the weight and the bias of the linear model (see Fig.\ref{fig:framework} for illustration). 
From eq.~\ref{eq:randfeat_mmepred}, we can express the output response of $f$ as the mean of all $s^{(k, i)}$, which we denote as \emph{the score} of the $i$-th cell in the $k$-th sample-set $\mathcal{X}^{(k)}$. This formulation naturally allows to quantify and interpret the contribution of every cell to the final prediction, which can potentially lead to an improved understanding of biological phenomena and enable better diagnoses and treatment for patients.\\

\begin{algorithm}[t!]
\caption{\textsc{Compute the Mean Embedding After Sub-selection Via Kernel Herding}}
\begin{algorithmic}[1]
\label{algo_concise}

  \REQUIRE A sample-set $\mathcal{X}$, number of cells kept after sub-selection $m$, dimensionality of the random feature space $D$, kernel hyperparameter $\gamma$.
          
\STATE \texttt{\# Compute Random Fourier Frequency Features}
\STATE Compute $\mathbf{W}\in \mathbb{R}^{d \times \frac{D}{2}}$ by sampling its elements independently $\mathbf{w}_{i,j} \sim \mathcal{N}(0, \frac{1}{\gamma})$ 
\FOR {each $x^{(i)} \in \mathcal{X}$}
    \STATE $\varphi({x}^{(i)}) \leftarrow [\text{sin}(\mathbf{W}^T x^{(i)}), \text{cos}(\mathbf{W}^T x^{(i)})] \in \mathbb{R}^D$, where $[\cdot,\cdot]$ denotes concatenation.
\ENDFOR

\STATE  \texttt{\# Sub-selection with Kernel Herding}

\STATE Initialize $j \leftarrow 1$,  $\mathcal{\hat{X}} \leftarrow \emptyset$, $\theta_0 \leftarrow \frac{1}{n}\sum_{i=1}^{n} \varphi(x^{(i)})$, $\theta \leftarrow \theta_0$

\WHILE{$j \leq m$}
    \STATE $i^* \leftarrow \underset{i}{\arg   \max} \ \theta^T \varphi(x^{(i)})$
    \STATE $\mathcal{\hat{X}} \leftarrow \mathcal{\hat{X}} \cup \{\varphi(x^{(i^*)})\}$,
    \STATE $\theta \leftarrow \theta + \theta_0 - \varphi(x^{(i^*)})$
    \STATE $j \leftarrow j + 1$
\ENDWHILE
\STATE \texttt{\# Compute the mean embedding of the Sub-selected Sample-set}
\STATE $\mu_{\mathcal{\hat{X}}} = \frac{1}{m}\sum_{z  \in \mathcal{\hat{X}}} z$
\STATE \textbf{return} $\mu_{\mathcal{\hat{X}}}$

\end{algorithmic}
\end{algorithm}

\noindent 

\noindent {\bf Kernel Herding}\ \ When using random features (eq.~\ref{eq:randfeat_mme}),  $\mu_\mathcal{X}$ may be understood as the average of random features for cells found in a sample-set. Predicted outputs based on  $\mu_\mathcal{X}$ (eq.~\ref{eq:randfeat_mmepred}) may be further interpreted as the average ``score'' of cells, $s^{(i)}$, in the respective sample-set. However, sample-sets may contain many (hundreds of) thousands of cells, making it cumbersome to analyze and synthesize the cell scores in a sample-set. To ease interpretability, we propose to sub-select cells in the sample-set in a way that yields a similar embedding to the original sample-set. That is, we wish to find a subset $\hat{\mathcal{X}} \subset \mathcal{X}$ such that $\mu_{\hat{\mathcal{X}}} \approx \mu_{\mathcal{X}}$, which implies that one may make similar inferences using a smaller (easier to interpret) subset of cells as with the original sample-set. Although a uniformly random subsample of $\mathcal{X}$ would provide a decent approximation $\mu_{\hat{\mathcal{X}}}$ for large enough cardinality ($|\hat{\mathcal{X}}|$), it is actually a suboptimal way of constructing an approximating subset \cite{chen2010super}. Instead, we propose to better construct synthesized subsets (especially for small cardinalities) using kernel herding (KH) \cite{chen2010super}. KH can provide a subset of $m$ points that approximates the mean embedding as well as $m^2$ uniformly sub-sampled points. We expound on KH for producing subsets of key predictive cells in Algorithm \ref{algo_concise}. We denote the dataset with the sub-selected sets as $\hat{\mathcal{D}}=\big\{(\hat{\mathcal{X}}^{(k)}, y^{(k)})\big\}_{k=1}^{N}$.\\\\

\section{Results}

\subsection{Datasets}
We used publicly available, multi-sample-set flow and mass cytometry (CyTOF) datasets. Each sample-set consists of several protein markers measured across individual cells. 

\textit{Preeclampsia.}
The preeclampsia CyTOF dataset profiles 11 women with preeclampsia and 12 healthy women throughout their pregnancies. Sample-sets corresponding to profiled samples from women are publicly available\footnote{\url{http://flowrepository.org/id/FR-FCM-ZYRQ}}. Our experiments focused on uncovering differences between healthy and preeclamptic women. 

\textit{HVTN.}
The HVTN (HIV Vaccine Trials Network) is a Flow Cytometry dataset that profiled T-cells across 96 samples that were each subjected to stimulation with either Gag or Env proteins \footnote{Note these are proteins meant to illicit functional responses in immune cell-types.} \cite{flowcap}. The data are publicly available\footnote{\url{http://flowrepository.org/id/FR-FCM-ZZZV}}. Our experiments focused on uncovering differences between Gag and Env stimulated samples.

\textit{COVID-19.} The COVID-19 dataset analyzes cytokine production by PBMCs derived from COVID-19 patients. The dataset profiles healthy patients, as well patients with moderate and severe covid cases. Specifically, the dataset consists of samples from 49 total individuals and is comprised of 6 healthy, 23 labeled intensive-care-unit (ICU) with moderately severe COVID, and 20 Ward (non-ICU, but covid-severe) labeled individuals with severe COVID cases, respectively. The data are publicly available\footnote{\url{http://flowrepository.org/id/FR-FCM-Z2KP}}. Our experiments focused on uncovering differences between sample-sets from ICU and Ward patients.

\begin{table}[t!]
\centering
\caption{Classification accuracies on the three datasets. Standard deviations are computed from 5 independent runs.}\label{tab:acc}
    \begin{tabular}{c|c|c|c}
    \toprule
          &  HVTN & Preeclampsia &  COVID-19  \\
    \midrule
        CytoDx \cite{cytoDX} & 65.24 $\pm$ 1.90 & 56.17 $\pm$ 3.95 & 68.82 $\pm$ 3.15 \\
        CellCNN \cite{cellCNN} & 81.87 $\pm$ 1.77 & 58.51 $\pm$ 3.25 & 75.93 $\pm$ 2.62 \\
        CytoSet \cite{yi2021cytoset} & \textbf{90.46 $\pm$ 2.20}  & 62.85 $\pm$ 3.42 & \textbf{86.55 $\pm$ 1.76} \\
    \midrule
        CKME & \textbf{90.68 $\pm$ 1.69} & \textbf{63.52 $\pm$ 2.22} & \textbf{86.38 $\pm$ 1.92} \\
    \bottomrule
    \end{tabular}
\end{table} 

\begin{table}[t!]
\centering
\caption{Number of parameters ($\times 10^3$) for different methods on different datasets.}\label{tab:num_params}
    \begin{tabular}{c|c|c|c}
    \toprule
          &  HVTN & Preeclampsia &  COVID-19  \\
    \midrule
        CytoDx \cite{cytoDX} & \textbf{0.01}  & \textbf{0.03} & \textbf{0.03} \\
        CellCNN \cite{cellCNN} & 40.7 & 20.2 & 40.5 \\
        CytoSet \cite{yi2021cytoset} & 330.2  & 80.0 & 330.0 \\
    \midrule
        CKME & 2.0 & 2.0 & 2.0 \\
    \bottomrule
    \end{tabular}
\end{table}

\subsection{Baselines}
We focus our comparison to current state-of-the-art gating-free methods CytoDx \cite{cytoDX}, CellCNN \cite{cellCNN}, CytoSet \cite{yi2021cytoset}, which look to assuage the drawbacks of gating-based methods. 
Please refer to Sec.~\ref{sec:intro}  ``Related Work'' for a detailed discussion about these methods. 


\subsection{Implementation Details}
Given the limited number of sample-sets, we used 5-fold cross-validation to report the classification accuracies on the three datasets. We tuned hyperparameters (e.g. the bandwidth parameter $\gamma$) on the validation set. On the HVTN and the Preeclampsia datasets, $\gamma$ was set to 1 while on the COVID-19 dataset $\gamma$ was set to 8. The dimensionality $D$ was set to 2000 for all datasets. The linear classifier in eq.~\ref{eq:randfeat_mmepred} is implemented by a linear SVM. 
For all the methods, $m=200$ cells are sub-selected for all datasets. The best hyperparameters of CytoSet, CellCNN and CytoDx are selected based on the validation performance. CytoSet, CellCNN, CytoDx was run using their public implementation\footnote{CytoSet: \url{https://github.com/CompCy-lab/cytoset}, CellCNN: \url{http://eiriniar.github.io/CellCnn/code.html}, CytoDx: \url{http://bioconductor.org/packages/CytoDx}}.

\begin{figure}[t!]
\begin{center}
  \includegraphics[width=0.8\columnwidth]{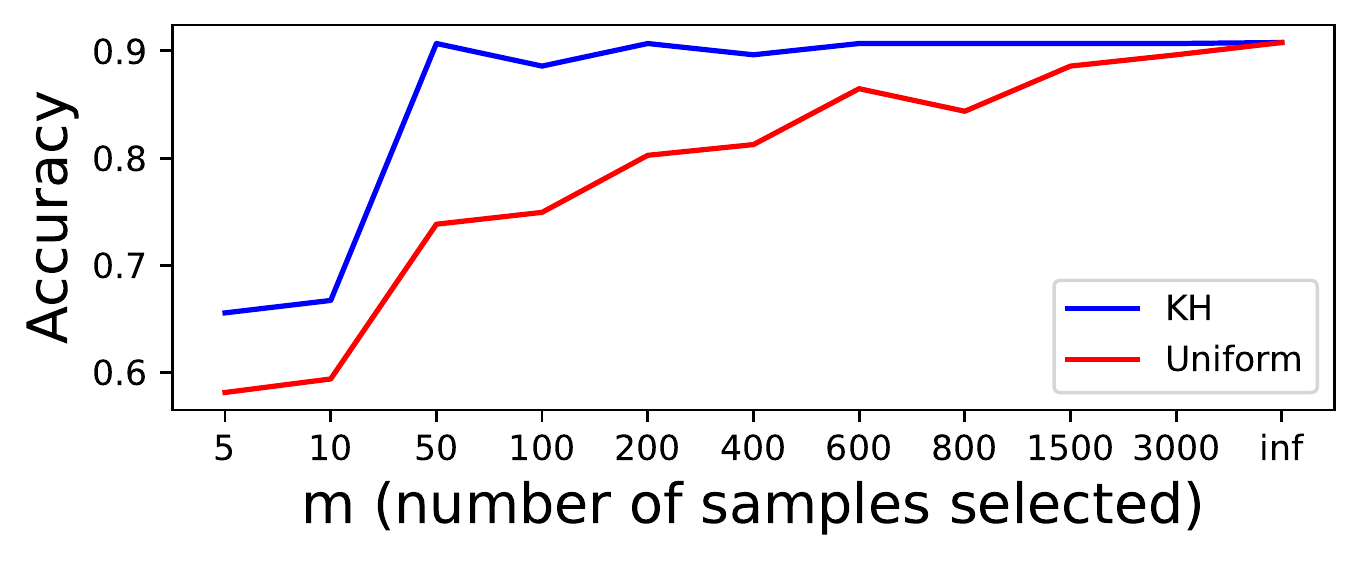}
\caption{The influence of the number of cells selected by KH and uniform sub-sampling, $m$, on the classification accuracy. $m=$``inf'' means no sub-selection.}
\label{fig:1}
\end{center}  
\end{figure}

\subsection{Classification Accuracies}
We report accuracies in Table \ref{tab:acc} and numbers of model parameters in Table \ref{tab:num_params}. Although our foremost goal is to make a more transparent model, we find that CKME achieves comparable or better accuracies than the state-of-the-art gating-free methods. 
The strength of CKME's performance is impressive when taking into account that:
1) CKME contains on average two orders of magnitude fewer parameters than CytoSet;
2) unlike CytoSet, which has an opaque, uninterpretable set-featurization and prediction model, CKME uses a simple average score over cells;
3) comparably simple models, such as CytoDX and CellCNN, have significantly poorer accuracies than CKME.
We believe that CKME's success is due to the expressive ability of random Fourier features, which provide flexible non-linear mappings (to cell scores) over input cell features without the need to learn additional featurization (e.g. from a neural network).
Moreover, the average of the random Fourier features, the mean map embedding (Sec.~\ref{sec:methods} ``Random Fourier Features''), is descriptive of the overall distribution and composition of respective sample-sets; since the discriminator is trained directly on the mean map embeddings, CKME is able to classify based on the cell composition of sample-sets.
Lastly, when a sample-set is sub-selected with kernel herding (Sec.~\ref{sec:methods} ``Kernel Herding''), the selected cells are explicitly a salient, descriptive subset that distills the sample-set whilst preserving overall characteristics.

\subsection{Ablation Studies}

\noindent \textbf{Ablation Studies on Sub-sampling}\ \ We first ablate the number of sub-samples to study the impact of cell sub-selection on accuracy. Fig. \ref{fig:1} shows the classification accuracies of CKME on the HVTN dataset with different numbers of cells selected by Kernel Herding (KH) and uniform sub-selection. We found that KH consistently outperforms uniform sub-selection and the accuracy of KH quickly saturates with as few as 50 cells sub-selected for every sample-set, confirming the strong capacity of Kernel Herding to maintain the distribution of the original sample-sets after sub-selection.
Moreover, in Table \ref{tab:ablation_KH}, we show the performance of CKME with 200 cells sub-selected by uniform sub-sampling (CKME w/ Unif). Compared to sub-selection with Kernel Herding, we find that using uniform sub-sampling leads to a worse result. This confirms that Kernel Herding is more effective than uniform sub-sampling.  \\

\begin{table}[t!]
    \centering
    \caption{Ablation Studies. }
    \label{tab:ablation_KH}
    \begin{tabular}{c|c|c|c}
    \toprule
       & HVTN & Preeclampsia & COVID-19 \\
    \midrule
        
        CKME w/ Unif & 80.26 $\pm$ 1.53  &  55.52 $\pm$ 4.33 & 84.28 $\pm$ 1.74 \\
        
        Naive Mean & 64.24 $\pm$ 3.17 & 57.60 $\pm$ 4.20 & 83.07 $\pm$ 2.29 \\
    \midrule
        CKME  & \textbf{90.68 $\pm$ 1.69} & \textbf{63.52 $\pm$ 2.22}  & \textbf{86.38 $\pm$ 1.92} \\
    \bottomrule
    \end{tabular}
\end{table}

\begin{figure*}[h]
         \begin{center}
         \includegraphics[ width=0.75\textwidth]{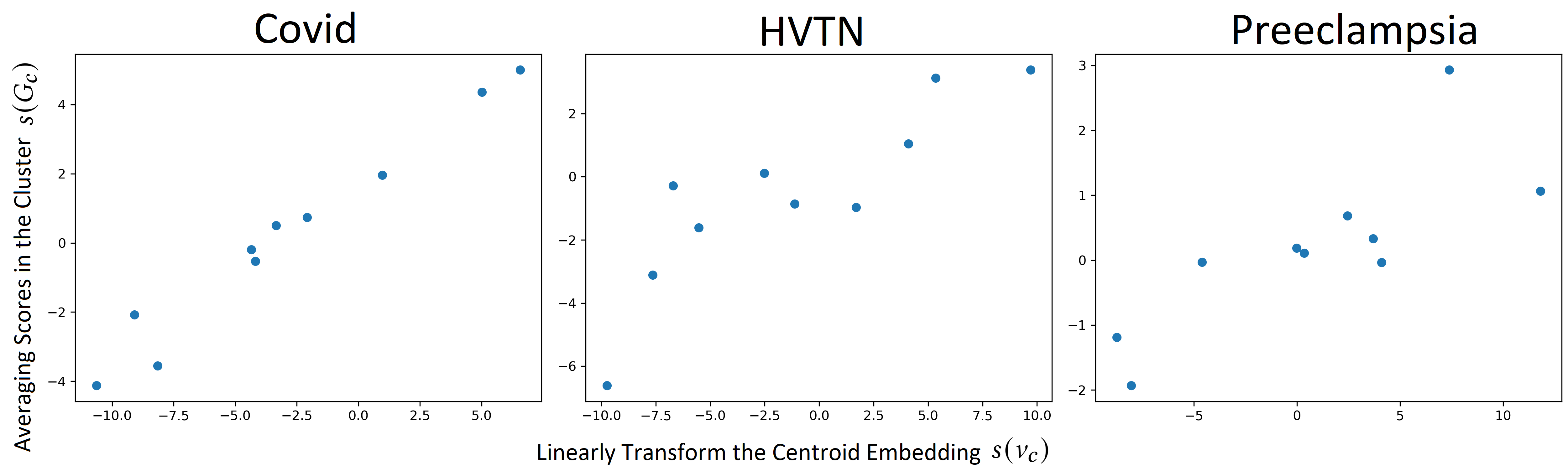}
         \caption{Cluster scores computed by the two methods are positively correlated.}
         \label{fig:corr2}
         \end{center} 
\end{figure*}

\noindent \textbf{Ablation Studies on Naive Mean Embedding}\ Here, we investigate the performance of mean embeddings without using random Fourier features. In this case, we still used Kernel Herding to sub-select $m$ cells but represented the summarized sample-set in the original feature space as $\bar{\mathbf{x}}^{(k)} = \frac{1}{m}\sum_{j=1}^{m}\hat{x}^{(k,j)} \in \mathbb{R}^d$. We call this approach ``Naive Mean''. As shown in Table \ref{tab:ablation_KH}, ``Naive Mean'' is worse than CKME, which indicates that simple summary statistics ($\bar{\mathbf{x}}^{(k)}$, the mean of input features) does not suffice for classification; in contrast, the random feature kernel mean embedding is able to provide an expressive enough summary of sample-sets.

\section{Analysis}

\subsection{Semantic Analysis of Cell Scores}
Recall that CKME can compute a score, $s^{(k,i)}$ (eq.~\ref{eq:randfeat_mmepred}), for every cell feature vector $\hat{x}^{(k,i)}$ in the KH sub-selected sample-set $\mathcal{\hat{X}}^{(k)}$.
Although this already enables our model to be more transparent, we hope that these scores are semantically meaningful and consistent with intuitive properties that one would want. 
We draw an analogy to natural language processing word-embeddings \cite{mikolov2013distributed}, where one constructs word-level featurizations (embeddings) that one hopes are semantically meaningful. For example, the sum of a group of consecutive words-embeddings in a sentence should yield a sentence-embedding that maintains the characteristics of that sentence as a whole.
Here, we study the semantics of scores of regions (groups) of cells induced by our cell scores. 

We may assign scores to regions in two ways: 1) averaging scores of cells in a nearby region ; 2) directly compute the score of the centroid of the respective region. That is, for cells in a nearby region, $G = \{x^{(i)}\}_{i=1}^m$, their average score is $s(G) = \frac{1}{m} \sum_{i=1}^m \varphi(x^{(i)})^T\beta + b$, where $\varphi: \mathbb{R}^d \mapsto \mathbb{R}^D$ are the Fourier features, and $\beta, b$ are the parameters of the learned linear model. In contrast, one may \emph{directly} compute the score of the centroid of this region, $\nu = \frac{1}{m} \sum_{i=1}^m x^{(i)}$, as $s(\nu) = \varphi(\nu)^T\beta + b$ (note the abuse of notation on $s$). Both $s(G)$ and  $s(\nu)$ score a region. Analogously to word-embeddings we hope that the score of cells (``words'') retains the semantics of the regions (``sentences''). Below we study the semantic retention of cell scores through a comparative analysis between $s(G)$ and $s(\nu)$, and a predictive analysis. \\

\begin{figure}[t]
\begin{center}
  \includegraphics[width=0.99\columnwidth]{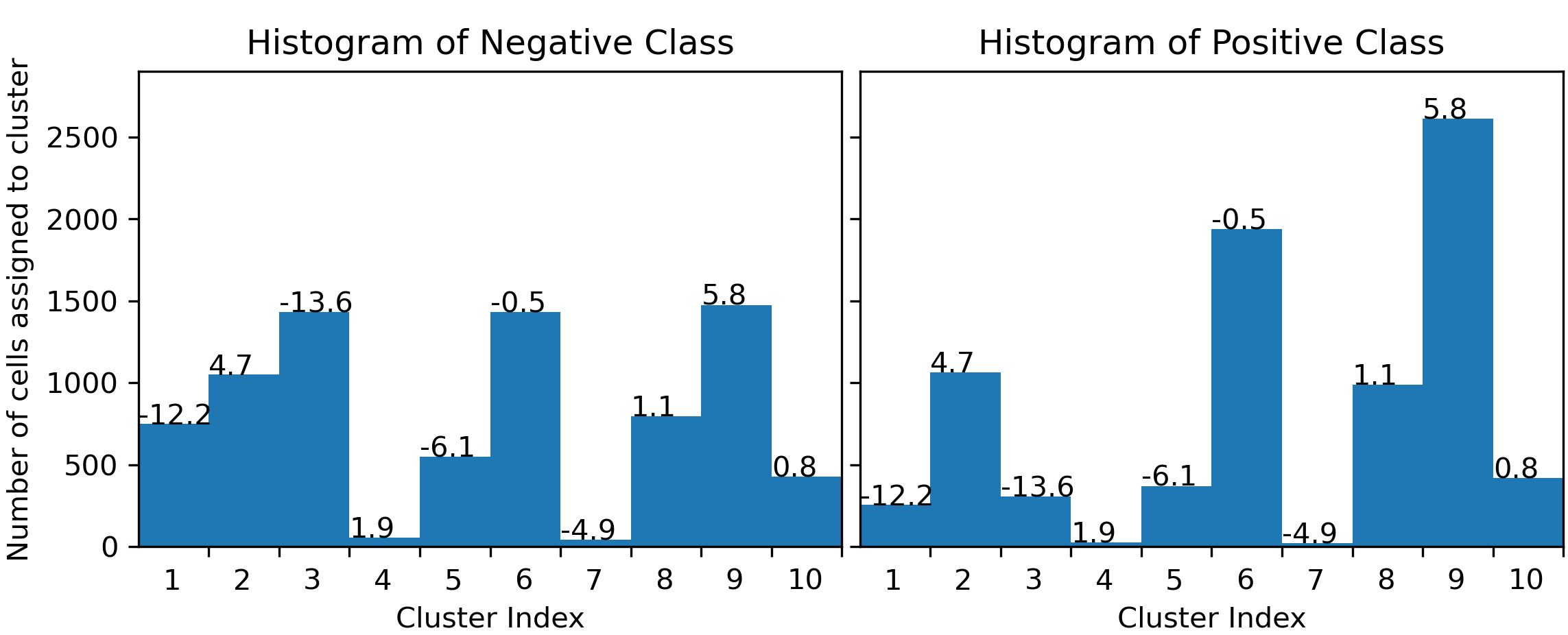}
\caption{The histogram of the clustering assignments on the HVTN dataset. The scores of all clusters computed by linearly transforming the corresponding centrioid embedding, $s(\nu_c)$, is shown on the top of each corresponding bar.}
\label{fig:2}
\end{center}  
\end{figure}

\noindent \textbf{Regions}\ \ We construct nearby regions with a k-means cluster analysis of cells in each dataset. I.e., we obtain $C$ cluster centroids, $\nu_1, \ldots, \nu_C$, and respective partitions $G_c = \{x \in \bigcup_{k=1}^N \mathcal{\hat{X}}^{(k)} \, |\, c = \mathrm{argmin}_l ||\nu_l - x||\}$.
Below we set $C=10$. The distribution of cluster assignments among cells from positive and negative classes on HVTN is shown in  Fig. \ref{fig:2}. One can see that clusters with a negative score (e.g.~Clusters 1 \& 3) tend to be more heavily represented in negative sample sets than in positive sample sets (and vice-versa for positive clusters such as Cluster 9).
\\

\noindent \textbf{Comparative Analysis}\ \ First, we analyze the semantic retention of scores by studying the correlation between the direct scores of cluster centroids $s(\nu_c) = \varphi(\nu_c)^T\beta + b$, to the average score of cells in respective clusters $s(G_c) = \frac{1}{|G_c|} \sum_{x \in G_c} \varphi(x)^T\beta + b$. \emph{Both $s(\nu_c)$ and $s(G_c)$ assign scores to the same region}, however, due to the nonlinearity of Fourier features there is no guarantee that $s(G)$ will match $s(\nu)$ as generally $\frac{1}{m} \sum_{i=1}^m \varphi(x^{(i)})^T\beta + b \neq \varphi\left(\frac{1}{m} \sum_{i=1}^m x^{(i)}\right)^T\beta + b$. 
Notwithstanding, if cell scores are semantically meaningful, then $s(\nu_c)$ and $s(G_c)$ should relate to each other as they both score regions. When observing the scatter plot of $s(\nu_c)$ vs. $s(G_c)$ across our datasets (see Fig.~\ref{fig:corr2}), we see that both scores on regions are heavily correlated; this shows that our scores are capable of semantically retaining the characteristics of regions. The respective Pearson correlation coefficients for COVID-19, HVTN, and Preeclampsia are 0.9785, 0.8234, and 0.7654.  \\

\begin{table}[t!]
\centering
\caption{Classification accuracies on the three datasets. Standard deviations are computed from 5 independent runs.}\label{tab:cluster_acc}
    \begin{tabular}{c|c|c|c}
    \toprule
          &  HVTN & Preeclampsia &  COVID-19  \\
    \midrule
         \textit{Linear} & 78.66 $\pm$ 2.86  & 56.64 $\pm$ 4.16  & 76.92 $\pm$ 2.05  \\
       
        \textit{Centroid CKME}  & 76.48 $\pm$ 2.62  & 60.57 $\pm$ 3.19 & 77.88 $\pm$ 2.11\\
        \textit{Average CKME} & 74.51 $\pm$ 2.46 & 60.80 $\pm$ 3.03 & 76.17 $\pm$ 2.13  \\
        \textit{Average MELD} & 64.52 $\pm$ 0.92 & 59.64 $\pm$ 3.47 & 59.16 $\pm$ 2.72 \\
        
    \bottomrule
    \end{tabular}
\end{table}

\noindent \textbf{Predictive Analysis}\ \ Next, we further study the semantic retention of cell scores with a predictive analysis. Traditionally \cite{citrus,spade,vopo} one may predict a label $y$ using a cluster analysis through a frequency-featurization. E.g. with a linear model $\hat{f}(\mathcal{\hat{X}}^{(k)}) = \sum_{c=1}^C \hat{r}^{(k)}_c \alpha_c + a$, where $\hat{r}^{(k)}_c$ is the proportion of cells in $\mathcal{\hat{X}}^{(k)}$ assigned to cluster $c$, and $\alpha, a$ are parameters of a learned linear model. In contrast to learning a linear model, one may directly build a predictor using scores for clusters. That is, it is intuitive to combine the scores of clusters using a convex combination according to the frequency of clusters in a sample-set: $\hat{f}_s(\mathcal{\hat{X}}^{(k)}) = \sum_{c=1}^C \hat{r}^{(k)}_c s_c$, where $s_c$ is the (pre-trained) score of the $c$-th region as described above, which is acting as a linear coefficient in the predictor $\hat{f}_s$. I.e. if we have scores for each cluster (coming from a pre-trained CKME model), then weighing these scores according to their respective prevalence in sample-sets should be predictive. Note that although this line of reasoning is \emph{semantically intuitive}, the CKME cell scores were not trained to be used in this fashion, hence there is no guarantee that $\hat{f}_s$ will be predictive.
We compare the accuracy of learning a linear model over cluster frequency features (denoted as \textit{Linear}), $\hat{f}$, to directly predicting using fixed cluster scores, ${f}_s$, in Table \ref{tab:cluster_acc}. We consider cluster scores coming directly from respective centroids $s_c = s(\nu_c) =  \varphi(\nu_c)^T\beta + b$ as \textit{Centroid CKME}, and cluster scores coming from averages $s_c = s(G_c) = \frac{1}{|G_c|} \sum_{x \in G_c} \varphi(x)^T\beta + b$ as \textit{Average CKME}. As an additional baseline, we also considered building a predictor through cell scores derived from MELD \cite{meld}. MELD is an alternative method that computes a score for every cell by first estimating the probability density of each sample along a $k$NN graph and then computes the relative likelihood of observing a cell in one experimental condition relative to the rest. As MELD cannot directly assign scores to post-hoc centroids, we utilize the average MELD scores for prediction $s_c = \frac{1}{|G_c|} \sum_{x \in G_c} s_x^{\mathrm{MELD}}$, denoted as \textit{Average MELD}. As shown in Table \ref{tab:cluster_acc}, predictors based on both CKME cluster scores (\textit{Centroid CKME} and \textit{Average CKME}) perform comparably to learning a linear model (\textit{Linear}). Here we see that these scores are semantically meaningful since they retain intuitive predictive properties after manipulations. Again, this is surprising given that our CKME cell scores were not trained for this purpose (for prediction with cluster frequencies). This point is further emphasized by the lesser predictive performance of scores from MELD (\textit{Average MELD}), which also seeks to provide semantically meaningful scores and was not trained for prediction in this fashion.

\begin{figure}[t]
\begin{center}
  \includegraphics[width=0.5\columnwidth]{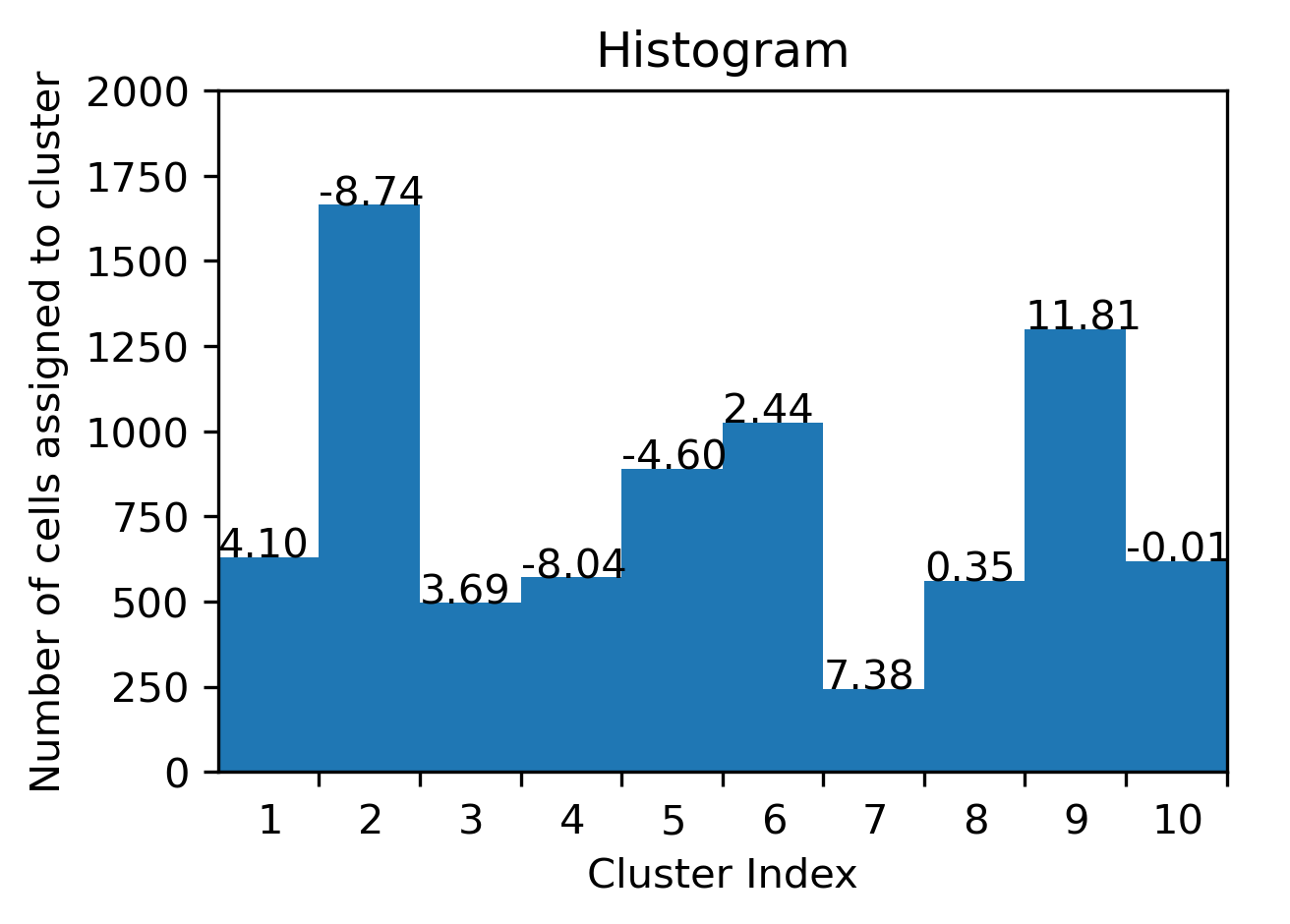}
\caption{We computed the frequency of (e.g. number of) cells assigned to each of 10 clusters, or cell-populations. We further computed the score of corresponding centroids (value shown above each bar). 
}
\label{fig:app1}
\end{center}  
\end{figure}

\begin{figure*}[h]
\begin{center}
  \includegraphics[width=2.0\columnwidth]{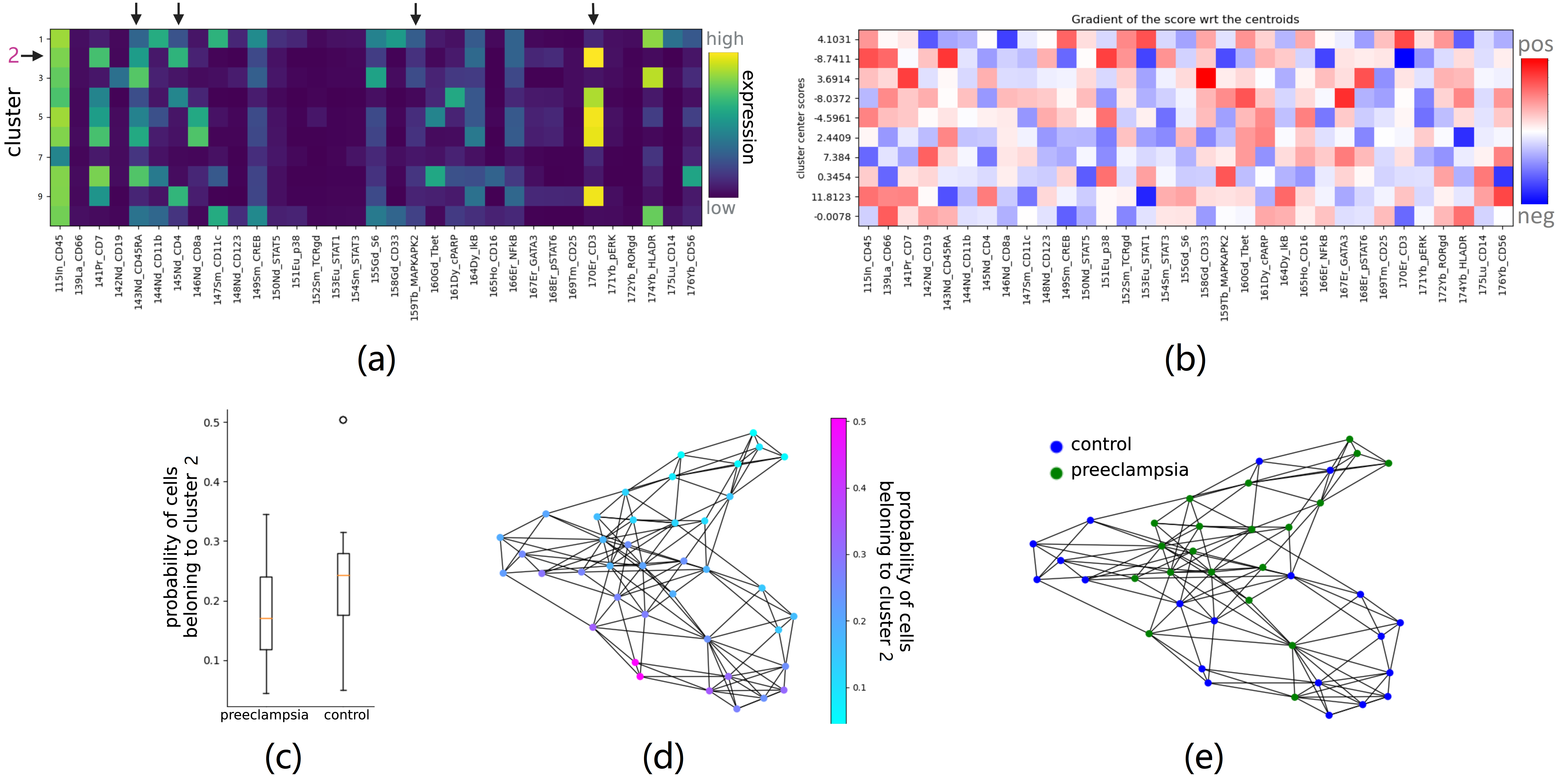}
\caption{Our predicted scores prioritized cluster 2 (CD4$^{+}$ naive T-cells) as cell-population likely to have frequency differences between control and preeclamptic samples. (a) Cluster 2 was identified to correspond to a population of MAPKAPK2$^{+}$ CD4$^{+}$ naive T-cells according to protein markers (denoted with arrows). (b) Gradients of CKME scores for cluster centroids $\nu_c$, $\nabla_x (\varphi(x)^T\beta)\,|_{x=\nu_c}$, where $\varphi(x)$ are random Fourier features and $\beta$ are the learned model weights (eq.~\ref{eq:randfeat_mmeout}). (c) The distributions of frequencies of cells assigned to cluster 2 in sample-sets from preeclamptic and control women. (d) A $k$-NN graph connecting samples-sets (nodes) according to computed frequencies across cell-populations reveals a higher frequency of cells assigned to cluster 2 in control samples-sets. (e) The $k$-NN graph from ({\bf d}), with each sample-set (node) colored by its ground-truth label.}
\label{fig:pree_interpret}
\end{center}  
\end{figure*}

\begin{figure*}[h]
\begin{center}
  \includegraphics[width=2.0\columnwidth]{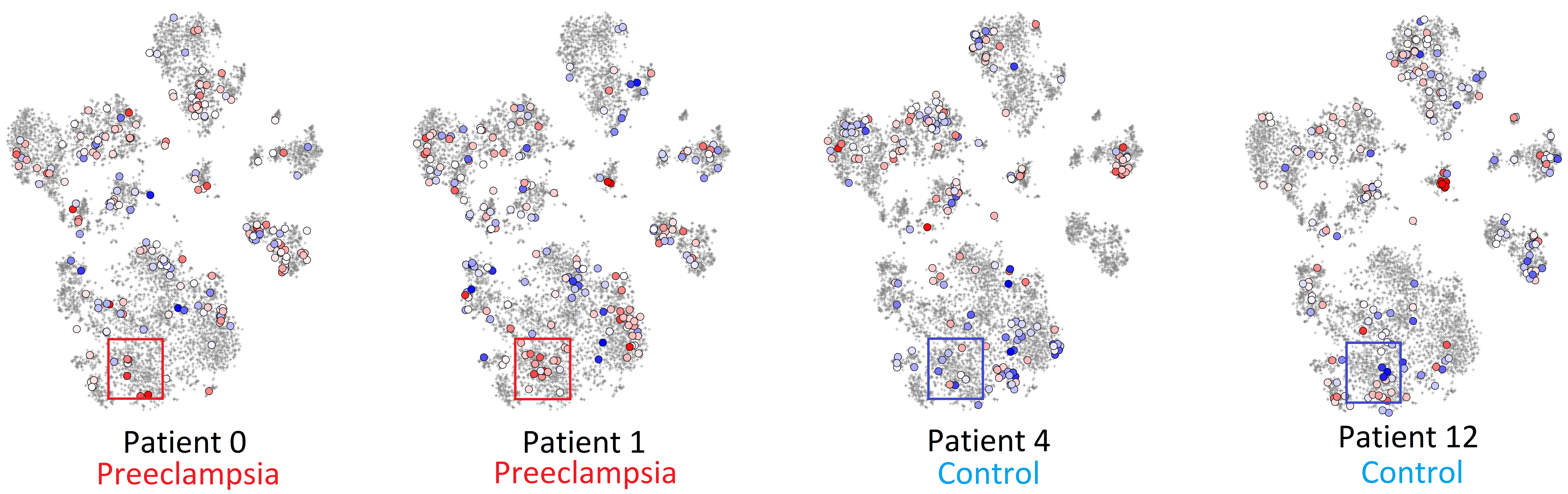}
\caption{Visualizing computed per-cell CKME scores via tSNE. Cells in gray represent a general, patient-wide cellular landscape. Cells colored red (blue) imply high (low) associations with the preeclampsia phenotype.}
\label{fig:pree_tsne}
\end{center}  
\end{figure*}

\subsection{Biological Validation} 

An important advantage of interpretable models is their capacity to uncover scientific knowledge \cite{carvalho2019machine}. Therefore, we would like to know whether the patterns automatically learned by CKME are consistent with existing knowledge discovered independently via manual analysis. To do so, we focus our analysis on a salient cluster in the Preeclampsia dataset. More specifically, we used our scores to prioritize cells that were different between control (healthy) and preeclamptic women. Cells across sample-sets were clustered into one of 10 clusters. We then computed the average score of the cells assigned to each cluster.  
We first identified cell-populations associated with patient phenotype using CKME scores for each cluster, followed by a finer per-cell analysis highlighting clinically-predictive individual cells and lastly identifying their defining features.

\subsubsection{Identifying Phenotype-Associated Cell-Populations} 
To link particular clusters (e.g. cell-populations) to clinical phenotype, we first prioritized cluster 2, based on its highly negative score according to CKME ($-8.74$, see Fig. \ref{fig:app1}). A highly negative score implies that it likely contained a large number of cells predicted as ``control''. The prominent protein markers expressed in cluster 2 were CD3, CD4, CD45RA, and MAPKAPK2 and indicated this is a cell-population of naive CD4$^{+}$ T cells expressing MAPKAPK2 (Fig. \ref{fig:pree_interpret}{\bf a}). The negative score of cluster 2 aligns with previous work, which showed that women with preeclampsia exhibit a decrease in MAPKAPK2$^{+}$ naive CD4$^{+}$ T-cells during the course of pregnancy, while healthy, women exhibit an increase \cite{pree}.

We first compared the distributions of frequencies of cells assigned to cluster 2 (i.e. this population of naive CD4$^{+}$ T cells expressing MAPKAPK2) between sample-sets from preeclamptic and healthy control women (Fig. \ref{fig:pree_interpret}{\bf c}). Consistent with our previous observations, sample-sets from control women had a statistically significantly higher proportion of cells assigned to cluster 2 in comparison to preeclamptic women (see Fig. \ref{fig:pree_interpret}{\bf c} with a p-value of $p=0.038$ under a Wilcoxon Rank Sum Test). As a complementary visualization, we constructed a $k$-nearest neighbor graph between sample-sets according to the computed frequencies across all cell-types (Fig. \ref{fig:pree_interpret}{\bf d-e}). Here, each node represents a sample-set and an edge represents sufficient similarity between a pair of sample-sets according to the frequencies of cells across cell-populations. In Fig. \ref{fig:pree_interpret}{\bf d}, sample-sets (nodes) are colored by the probability of their cells belonging to cluster 2. In comparison to sample-sets (nodes) colored by their ground-truth labels (Fig. \ref{fig:pree_interpret}{\bf e}), we observed that control, healthy sample-sets such as the densely connected set of blue nodes in the bottom of Fig. \ref{fig:pree_interpret}{\bf e} tend to have high frequencies of cells assigned to cluster 2.

\subsubsection{tSNE Visualizations of Per-Cell CKME Scores in Patient Samples in the Preeclampsia CyTOF Dataset}
\begin{figure}[t!]
\begin{center}
  \includegraphics[width=0.75\columnwidth]{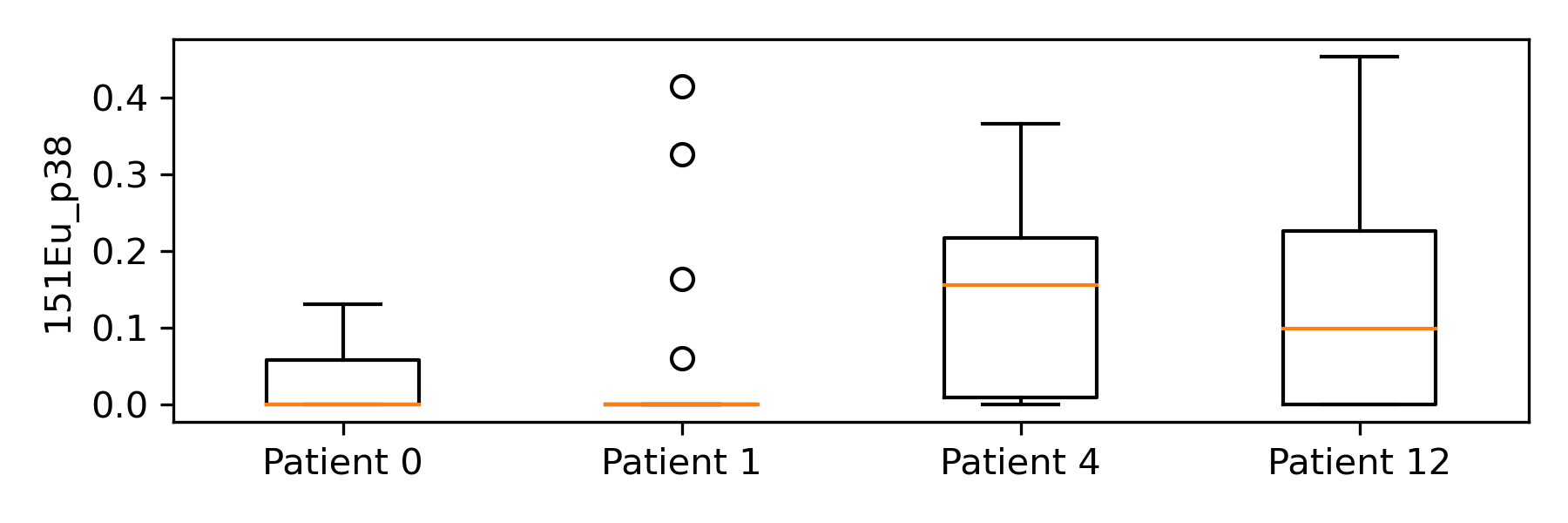}
\caption{Distributions of features p38 for the cells inside the rectangles of Fig. \ref{fig:pree_tsne}.}
\label{fig:p38}
\end{center}  
\end{figure}

Next, we perform a fine-grained analysis of individual cell scores.
In contrast to previous approaches to identifying phenotype-associated cell-populations on a population level \cite{vopo,citrus}, CKME provides much finer resolution and instead is able to highlight individual cells (in a digestible, synthesized KH subset) that are likely driving particular clinical phenotypes. For example, by combining cells from samples collected from healthy and preeclamptic women in the preeclampsia CyTOF dataset, we closely examined the patterns in the computed per-cell CKME scores and how they related to patient phenotypes. Briefly, 200 cells were sub-selected with Kernel Herding from each patient and the sub-selected cells from all patients are together projected into two dimensions with tSNE \cite{van2008visualizing} to establish a general, patient-wide cellular landscape (gray cells in Fig. \ref{fig:pree_tsne}). For each patient, their sub-selected cells in the two-dimensional space are colored by their computed CKME score (Fig. \ref{fig:pree_tsne}). In particular, a cell colored red (blue) implies a high (low) association with the
preeclampsia phenotype.
In Fig. \ref{fig:pree_tsne}, we visualized scores for patients sampled from two preeclamptic patients (Patients 0 and 1) and two control patients (Patients 12 and 4)\footnote{These patients were from a testing set and were not used to learn the CKME scores.}. Remarkably, in the two preeclamptic women, we identified prominent subsets of red-colored nodes (outlined in red rectangles), implying a strong association with the preeclampsia phenotype. Based on the expression of phenotypic markers, these cells were identified as memory CD4+ T-cells (based on the expression of CD3, CD4, and CD45RA). In contrast, the memory CD4+ T-cells in the healthy patients exhibited low scores (blue-colored cells outlined in blue rectangles), indicating their negative association with the healthy phenotype. 
Taken together, the CKME scores highlighted the memory CD4+ T-cells as a key predictive population (also shown in previous work \cite{pree}).
To further unravel the association between specialized immune cells and the preeclamptic and control patient phenotypes according to CKME scores, we investigated the prominent protein feature co-expression patterns of the subset of cells outlined in the rectangles in Fig. \ref{fig:pree_tsne}.  Comparing the two preeclamptic (patients 0 and 1) to the two control (patients 4 and 12) sample-sets (Fig. \ref{fig:pree_tsne}), we quickly identified further nuanced cellular subsets associated with the cells with high (red) and low (blue) scoring CKME scores. A particular differentiating protein marker between the preeclamptic and control patients in this memory CD4$^{+}$ T-cell population was p38. The distribution of its expression in cells in the rectangular regions is shown in Fig. \ref{fig:p38} and reveals prominent differences between the preeclamptic and control patients. p38 is a functional protein marker, indicating likely differences in signaling responses between control and preeclamptic samples.  Interestingly, previous work  \cite{pree} also independently showed that the expression of p38 in memory CD4$^{+}$ T-cells was an important feature for predicting preeclampsia status.

\subsubsection{Assessing Protein Features in Preeclamptic and Healthy Patients}
\label{sec:profeats}
Lastly, we investigated the protein feature coexpression patterns of immune cell types from preeclamptic and healthy patients to further determine whether CKME scores are consistent with existing knowledge and could be used for hypothesis generation. To identify the cell populations associated with extremely positive or negative average CKME scores, we further annotated the clusters according to the expression of known protein markers (Fig. \ref{fig:pree_interpret}{\bf a}). We found that the cell types that were strongly associated with the preeclampsia phenotype (positive average CKME score) were clusters 1 and 9, corresponding to classical monocytes and memory CD4$^{+}$ T cells (Table \ref{tab:gradient}). In contrast, the cell types associated with the control phenotype (negative average CKME score) were clusters 2, 4, and 5 corresponding to naive CD4$^{+}$ T cells expressing MAPKAPK2, naive CD4$^{+}$ T cells, and naive CD8$^{+}$ T cells (Table \ref{tab:gradient}). Notably, this is consistent with previous work that found that the absolute monocyte count was significantly higher in preeclampsia patients than in controls \cite{Wang2019-jm,Vishnyakova2019-an}. Moreover as highlighted previously, an abundance of naive CD4$^{+}$ T cells (Fig. \ref{fig:pree_interpret}{\bf c-e}) or memory CD4$^{+}$ T cells (Fig. \ref{fig:pree_tsne}) are strong indicators of healthy or preeclampsia status, respectively \cite{Chaiworapongsa2002-ym,Darmochwal-Kolarz2007-ia}.

We also provide an alternative investigation of important protein markers in CKME scores. In particular, we study how minuscule changes in the expression of the protein markers defining the cluster (cell type) would increase the CKME score and thus have a stronger association with the clinical phenotype. To do so, we computed the gradient of the CKME score with respect to the cluster centroid. I.e.~we compute $\nabla_x (\varphi(x)^T\beta)\,|_{x=\nu_c}$ for cluster centroids $\nu_c$, , where $\varphi(x)$ are random Fourier features and $\beta$ are the learned model weights (eq.~\ref{eq:randfeat_mmeout}). Doing so will uncover what small changes in protein markers alter the CKME of cluster centers, which serves as one proxy for feature importance.

The gradient heatmap (as shown in Fig. \ref{fig:pree_interpret}{\bf b}) provides a value for each feature within a centroid, where the corresponding direction of change specifies whether an increase (positive) or decrease (negative) in protein expression would increase the association with the preeclampsia phenotype. By prioritizing large magnitude changes in particular features that define cell types, in addition to signaling markers, we observed a shift in the expression of prioritized proteins. When examining the gradient feature changes in cluster 1 (generally defined as classical monocytes by the expression of CD14$^{+}$ CD16$^{-}$), we observed a decrease in CD14 and an increase in CD16 expression (Fig. \ref{fig:pree_interpret}{\bf a-b}, Table \ref{tab:gradient}). This indicates that a transition to a nonclassical monocyte phenotype (CD14$^{\mathrm{med}}$ CD16$^{\mathrm{med}}$) may cause a sample-set to appear more preeclamptic. Interestingly, previous work has indicated that the subpopulation composition of monocytes (classical, intermediate, nonclassical) significantly varies between preeclamptic and control patients \cite{Vishnyakova2019-an}. With respect to cluster 2 (naive CD4$^{+}$ T cell population expressing MAPKAPK2), we observed a decrease in MAPKAPK2 and an increase in p38 expression (Fig. \ref{fig:pree_interpret}{\bf a-b}, Table \ref{tab:gradient}). This 
further highlights that p38 may be a good functional marker for distinguishing between healthy and preeclamptic patients in both naïve and memory CD4$^{+}$ T cells. 

\renewcommand\theadfont{}
\begin{table}[t!]
  \centering
  \caption{Gradient analysis of each cell-population highlights prominent features in clinically-associated cell-populations.}\label{tab:gradient}
  \scalebox{0.5}{
  \begin{tabular}{c c c c}
    \toprule 
    \thead{ID/CKME score} & \thead{cell type \\ (expression)} & \thead{gradient} & \thead{cell type \\ (gradient)} \\ 
    \midrule
        1/4.1031  & classical monocytes & $\downarrow$ CD14 $\uparrow$ CD16 $\uparrow$ STAT1 & nonclassical monocytes STAT1$^{+}$ \\
        9/11.8123  & memory CD4$^{+}$ T cells & $\uparrow$ CD4 $\downarrow$ CD45RA $\uparrow$ STAT3 & memory CD4$^{+}$ STAT3$^{+}$ T cells \\
    \midrule
        2/-8.7411 & naive CD4$^{+}$ MAPKAPK2$^{+}$ T cells & $\uparrow$ CD45RA $\downarrow$ MAPKAPK2 $\uparrow$ p38 & naive CD4$^{+}$ p38$^{+}$ T cells \\
        
        4/-8.0372 & naive CD4$^{+}$ T cells & $\uparrow$ Tbet $\uparrow$ GATA3 & CD4$^{+}$ Tbet$^{+}$ GATA3$^{+}$ T cells \\
        
        5/-4.5961 & naive CD8$^{+}$ Tbet$^{+}$ T cells & $\uparrow$ Tbet $\uparrow$ p38 & naive CD8$^{+}$ Tbet$^{+}$ p38$^{+}$ T cells \\
         
   \bottomrule
 \end{tabular}
 }
\end{table}

\section{Discussion and Conclusion}
In this work, we introduced CKME (Cell Kernel Mean Embedding) as a method to link cellular heterogeneity in the immune system to clinical or external variables of interest, while simultaneously facilitating biological interpretability. As high-throughput single-cell immune profiling techniques are being readily applied in clinical settings \cite{ganio,pree,ramin}, and there are critical needs to 1) accurately diagnose or predict a patient's future clinical outcome and 2) to explain the particular cell-types driving these predictions. Recent bioinformatic approaches have uncovered \cite{citrus,vopo,cytoDX} or learned \cite{cellCNN,yi2021cytoset} immunological features that can accurately predict a patient's clinical outcome.
Unfortunately, these existing methods must often make a compromise between interpretability (e.g.~ simpler, more transparent methods, such as \cite{cytoDX, cellCNN} and accuracy (e.g.~more complicated, opaque methods, such as \cite{yi2021cytoset}).
Our experimental results show that CKME allows for the best of both worlds, yielding state-of-the-art accuracies, with a simple, more transparent model. 

We leveraged the transparency of CKME with a thorough analysis.
First, we showed that cell scores stemming from CKME retain intuitive desirable properties 
for cell contributions. We studied the semantics of CKME cell scores in tasks that CKME \emph{was not explicitly trained for} and found that CKME cell scores were useful for those scenarios. This suggests that, through simple supervised training, CKME can be used to obtain insights in a myriad of analyses that it was not trained for.
Furthermore, cell-population, individual-cell, and protein feature coexpression analyses
yielded insights that aligned with previous literature (e.g.~\cite{pree}). This suggests that CKME is a robust approach for uncovering scientific knowledge.  


In summary, CKME enables a more comprehensive, automated analysis and interpretation of multi-patient flow and mass cytometry datasets and will accelerate the understanding of how immunological dysregulation affects particular clinical phenotypes. Future directions may include alternate synthesizing approaches to Kernel Herding and variants that weigh cells differently. Furthermore, in addition to the gradient-based analysis presented in Sec.~\ref{sec:profeats}, we shall explore other methodologies for assessing feature importance in CKME scores such as fitting local, sparse models \cite{lime}.

\begin{acks}
  This research was partly funded by NSF grant IIS2133595 and by NIH 1R01AA02687901A1.
\end{acks}

\bibliographystyle{ACM-Reference-Format}
\bibliography{sample-authordraft}




\end{document}